\documentclass[final,3p,times,twocolumn]{elsarticle}
\usepackage{amssymb}
\usepackage{xcolor,soul,framed} 
\colorlet{shadecolor}{yellow}
\usepackage{graphicx}
\usepackage[cmex10]{amsmath}
\usepackage{array}
\usepackage{multirow,textcomp,booktabs}
\usepackage[caption=false,font=normalsize,labelfont=sf,textfont=sf]{subfig}
\usepackage{textcomp}
\usepackage{stfloats} 
\usepackage{cite}
\usepackage{url}
\usepackage{algorithm}
\usepackage{algpseudocode}
\usepackage{bm} 
\usepackage{amssymb}
\usepackage{hyperref} 
\usepackage{setspace}
\usepackage{caption}
\usepackage{lineno}
\begin{document}
\begin{frontmatter}
\title{SemiDDM-Weather: A Semi-supervised Learning Framework for All-in-one Adverse Weather Removal}
\cortext[mycorrespondingauthor]{Corresponding author}
\author[inst1]{Fang~Long}
\ead{longfang@e.gzhu.edu.cn}
\author[inst1]{Wenkang~Su\corref{mycorrespondingauthor}}
\ead{swk1004@gzhu.edu.cn}
\author[inst1]{Zixuan~Li}
\ead{lizx@e.gzhu.edu.cn}
\author[inst2]{Lei~Cai}
\ead{lcai\_gxy@hqu.edu.cn}
\author[inst1]{Mingjie~Li}
\ead{limingjie@gzhu.edu.cn}
\author[inst1]{Yuan-Gen Wang\corref{mycorrespondingauthor}}
\ead{wangyg@gzhu.edu.cn}
\author[inst3]{Xiaochun~Cao}
\ead{caoxiaochun@mail.sysu.edu.cn}

\affiliation[inst1]{organization={School of Computer Science and Cyber Engineering},%Department and Organization
            addressline={Guangzhou University}, 
            city={Guangzhou},
           postcode={510006}, 
            state={Guangdong},
            country={China}
            }
\affiliation[inst2]{organization={College of Engineering},
            addressline={Huaqiao University}, 
            city={Quanzhou},
           postcode={361021}, 
            state={Fujian},
            country={China}
            }
\affiliation[inst3]{organization={School of Cyber Science and Technology},
            addressline={Shenzhen Campus of Sun Yat-sen University}, 
            city={Shenzhen},
           postcode={518107}, 
            state={Guangdong},
            country={China}
            }

\begin{abstract}
{
Adverse weather removal aims to restore clear vision under adverse weather conditions. Existing methods are mostly tailored for specific weather types and rely heavily on extensive labeled data. In dealing with these two limitations, this paper presents a pioneering semi-supervised all-in-one adverse weather removal framework built on the teacher-student network with a Denoising Diffusion Model (DDM) as the backbone, termed SemiDDM-Weather. As for the design of DDM backbone in our SemiDDM-Weather, we adopt the SOTA Wavelet Diffusion Model-Wavediff with customized inputs and loss functions, devoted to facilitating the learning of many-to-one mapping distributions for efficient all-in-one adverse weather removal with limited label data. To mitigate the risk of misleading model training due to potentially inaccurate pseudo-labels generated by the teacher network in semi-supervised learning, we introduce quality assessment and content consistency constraints to screen the ``optimal'' outputs from the teacher network as the pseudo-labels, thus more effectively guiding the student network training with unlabeled data. Experimental results show that on both synthetic and real-world datasets, our SemiDDM-Weather consistently delivers high visual quality and superior adverse weather removal, even when compared to fully supervised competitors. 
Our code and pre-trained model are available at~\href{https://github.com/longfafffa/SemiDDM-Weather}{this repository}.}
\end{abstract}


\end{frontmatter}



% \maketitle
\section{Introduction}
Images captured under adverse weather conditions such as haze, rain, snow, and adhering raindrops often suffer from degraded visual quality, which not only affects the aesthetics of the image, but also potentially jeopardizes the performance of various downstream vision applications (e.g., detection~\citep{shi2023fixated}, segmentation~\citep{xu2023pidnet}, and depth estimation~\citep{uy2023scade}). Consequently, removing the impact of adverse weather conditions is crucial and has drawn much attention to computer vision. 

Over the past decade, remarkable achievements in the visual quality of restored images for different adverse weather conditions have been made~\citep{cai2021joint, zhang2022beyond, ZHAO2024106428}. Most of them, however, are tailored to address a specific weather degradation, making it difficult to meet the demands in practical situations, as degradation usually varies in the wild. In this regard, some advanced \emph{all-in-one} methods~\citep{chen2022learning,ozdenizci2023restoring,liu2024residual} have been developed, capable of restoring clear images from various adverse weather conditions without the need for any other targeted designs. Despite their notable progress, all of them are designed to learn the many-to-one mapping in a supervised manner, which requires large amounts of pairwise degraded-clear image samples for training. However, acquiring such paired images under adverse weather conditions poses a considerable challenge. Semi-supervised learning emerges as a viable solution to overcome the scarcity of paired data across various computer vision applications. Yet, to our knowledge, the exploration of semi-supervised learning techniques for all-in-one adverse weather removal remains an open issue.

Adverse weather removal is widely recognized as an inherently ill-posed inverse problem, seeking to generate a plausible reconstruction for the degraded input. Recent works~\citep{ozdenizci2023restoring,li2019heavy} have reframed this challenge as a conditional generative modeling task aimed at generating perceptually more plausible samples by drawing from the posterior distribution. Denoising Diffusion Models (DDMs) have emerged as a potent solution, demonstrating remarkable abilities in learning the underlying data distribution~\citep{gao2023implicit, Lugmayr_2022_CVPR} and thus, hold great promise for this application. They are particularly adept at handling the complex many-to-one mappings that characterize adverse weather removal tasks. WeatherDiff~\citep{ozdenizci2023restoring} is a pioneer in employing DDMs for all-in-one adverse weather removal. However, its reliance on supervised training using large amounts of paired data, as well as its lengthy inference times, greatly limits its practicality.

To tackle the challenges presented, we propose a semi-supervised framework called \textbf{SemiDDM-Weather}, which stands out as the first to achieve consistently visual high-quality image restoration in all-in-one adverse weather removal using limited labeled data. The proposed SemiDDM-Weather adopts a teacher-student network~\citep{tarvainen2017mean} as the basis for semi-supervised learning, which consists of an identically structured network of student and teacher. The student network is trained using labeled and unlabeled images, while the teacher network is incrementally refined through an Exponential Moving Average (EMA) of the student network's parameters, devoted to generating pseudo-labels for the unlabeled data to reverse-guide the students' training. 
{Inspired by previous methods, we also approach it as a conditional generative modeling task by sampling from the posterior distribution, but here we are in a semi-supervised manner. Therefore, the model backbone must possess strong generative capabilities to learn many-to-one mapping relationships to effectively address the problem of removing various adverse weather conditions with a unified model under limited labeled data.}
Excitingly, diffusion models are well-suited for this task due to their superior ability to capture underlying data distributions, as well as their prowess in generating high-quality outputs. Nevertheless, commonly used diffusion models necessitate numerous sampling steps, hindering inference speed and potentially accumulating errors, making semi-supervised learning more challenging. In response to this defect, we customize a wavelet diffusion model to serve as the backbone of both the teacher and student networks, which is built upon the SOTA WaveDiff~\citep{phung2023wavelet}, by modifying its model input and loss function to better adapt to our adverse weather removal task. It requires only four sampling steps instead of the usual hundreds, significantly enhancing inference efficiency and thus reducing cumulative errors from incorrect pseudo-labels.
Specifically, we implement a crop-based training approach that processes $64\times64$ patches of the degraded image as inputs, optimizing computational efficiency without compromising training effectiveness. Furthermore, we refine the content consistency of the outputs relative to the conditional inputs by integrating perceptual loss into the generator's training and enhancing the original $\mathcal{L}_1$ loss with a hybrid of $\mathcal{L}_1$ and $\mathcal{L}_2$ losses, which leverages the precision of $\mathcal{L}_1$ loss while benefiting from the noise resilience of $\mathcal{L}_2$ loss to improve content restoration. Notably, since wavelets are highly frequency-aware, the introduction of the wavelet-based diffusion scheme in our SemiDDM-Weather further sharpens the perceptual quality of the restored images.

Considering the teacher network outputs do not always surpass those of the student network, there is a risk that potentially inaccurate pseudo-labels generated by the teacher network could mislead the training of the student network. To mitigate this risk, the targeted design for the pseudo-label to improve its accuracy has been made.
Specifically, rather than using the current outputs of the teacher network, here we use the ``optimal'' outputs generated by the teacher network during training as the pseudo-labels to guide the student network training with unlabeled data. As for how to choose the ``optimal'' ones, we not only incorporate the quality assessment but also introduce content consistency constraints between the generated images and the degraded ones. 

In light of the difficulties in acquiring Ground Truth (GT) under adverse weather conditions, traditional Full-Reference Image Quality Assessment (FR-IQA) metrics like PSNR and SSIM fall short in assessing the quality of image restoration. {Moreover, our method prioritizes the restoration of perceptually clearer images. Therefore, we did not use pixel-level metrics such as PSNR and SSIM, as these metrics do not fully capture perceptual image quality as perceived by humans, as shown in~\citep{Yu_2024_CVPR}. Instead, we used No-Reference Image Quality Assessment (NR-IQA) metrics for quality evaluation, which aligns with recent studies such as~\citep{wang2024selfpromer,zhang2024diffrestorerunleashingvisualprompts}. }Experimental results indicate that our approach markedly exceeds other SOTA fully-supervised methods, including those tailored for specific and all-in-one weather conditions. Additionally, our method not only counteracts the negative impact of adverse weather in downstream tasks but also outperforms other fully-supervised techniques in mitigating such effects.

In summary, our contributions are as follows:
\begin{itemize}

    \item {We present a pioneering semi-supervised all-in-one adverse weather removal framework and modify the SOTA Wavediff to serve as the backbone, making it possible for a unified model to effectively remove various adverse weather conditions with limited labeled data. 
    }

    \item We use the ``optimal'' outputs from the teacher network in terms of both quality and content, rather than its current outputs, as pseudo-labels, which facilitates more accurate pseudo-labels acquisition, thus benefiting the model training
    
    \item Owing to the introduction of wavelet in the model, its strong frequency awareness allows it to restore perceptually clearer images, even beyond the ground truth (GT).
    
    \item Though our approach is semi-supervised, experimental results validate that it outperforms the current SOTA fully-supervised ones in restoration capability for various adverse weather scenarios. 
\end{itemize}
    
\section{Related Work}
\subsection {Restoration in Adverse Weather Conditions}
Adverse weather removal has been extensively explored in recent years, and existing methods can be divided into two families, i.e., weather-specific methods and all-in-one methods. 

\textbf{Weather-Specific Methods.}
These methods are designed to address specific adverse weather degradations such as deraining, desnowing, raindrop removal, and among others. For \textit{rain streak removal}, 
Cai et al.~\citep{cai2021joint} utilized the depth and density information to restore images for rain conditions. Zhao et al.~\citep{ZHAO2024106428} developed a novel cycle contrastive adversarial framework incorporating Cycle Contrastive Learning (CCL) and Location Contrastive Learning (LCL) to address unsupervised single image deraining problems. Zhang et al.~\citep{zhang2022beyond} introduced a Paired Rain Removal Network (PRRNet), which leverages both stereo images and semantic information to remove rain streaks effectively.
For \textit{snow removal}, Wang et al.~\citep{wang2020dcsfn} adopted a multi-sub-networks structure to learn features with different scales to handle snow removal problems. Chen et al.~\citep{chen2023uncertainty} developed an innovative and efficient paradigm to improve the performance and generalization ability of snow removal by integrating the degradation perception and background modeling. Chen et al.~\citep{CHEN2023896} proposed an extremely lightweight recursive network (XLRNet) designed to address de-snowing problems, which is constructed using a single recursive strategy and two novel lightweight modules.
For \textit{raindrop removal}, Quan et al.~\citep{quan2019deep} devised a convolutional neural network (CNN) mechanism for raindrop removal. 
Wang et al.~\citep{WangSS21} introduce a context-enhanced representation learning and deraining network, with a novel two-branch encoder design to solve the raindrop removal problem.
Chen et al.~\citep{chen2023sparse} approached this problem from a global perspective to learn and model degradation relationships, thereby removing raindrop degradation.

Although the above methods show excellent removal capabilities for targeted weather conditions, they are difficult to apply to other types of weather conditions, which may cause a significant performance bottleneck in practical applications.

\textbf{All-in-One Methods.}
Regarding the above issue, recent researches~\citep{li2020all,valanarasu2022transweather,liu2024residual} have shifted towards unified (a.k.a., all-in-one) frameworks that aim to handle various types of weather removal using the same model. Li et al.~\citep{li2020all} designed a generator
with multiple task-specific encoders and a common decoder to handle various adverse weather degradations. Valanarasu et al.~\citep{valanarasu2022transweather} constructed an end-to-end Transformer-based framework to remove various adverse weather. Zhu et al.~\citep{zhu2023learning} incorporated a two-stage training strategy to explore both general weather features and specific weather-related features to solve this problem.
Liu et al.~\citep{liu2024residual} introduced a multi-patch skip-forward architecture for the encoder, facilitating the transfer of fine-grained features from the initial layers to deeper layers, thereby enhancing feature embedding with detailed semantics for adverse weather removal.
Recently, the emergence of DDMs with their strong capability in learning underlying data distributions and generating high-quality outputs has led to their application in this field. Notably, \"{O}zdenizci et al.~\citep{ozdenizci2023restoring} developed a method utilizing patch-based DDM for all-in-one adverse weather removal, achieving a significant advancement. However, although this approach presented commendable restoration results, it requires large amounts of sampling steps and labeled data, which limits its practicality.

\subsection{Semi-Supervised Learning}
When labeled data is limited, Semi-Supervised Learning (SSL) plays an important role in improving the model’s performance by leveraging unlabeled data. Pseudo-labeling is one of the most widely used SSL techniques~\citep{xie2020self,berthelot2019mixmatch}, which generates pseudo-labels for unlabeled data based on model predictions. Another technique is consistency regularization~\citep{rasmus2015semi,bachman2014learning}, which aims to effectively enhance model performance by enforcing consistency among the distributions of multiple augmented samples generated using different data augmentation techniques for the same unlabeled sample. For pseudo-labeling techniques, most algorithms typically employ the teacher-student network, where the teacher network is used to generate pseudo-labels for guiding the training of the student network with unlabeled data~\citep{tang2021humble,huang2023contrastive}. 
However, these approaches can be problematic, as the teacher's outputs aren't always better than the student's, potentially leading to misleading training with {incorrect} labels. Additionally, for consistency regularization techniques, most methods rely on the $\mathcal{L}_{{1}}$ loss for consistency regularization constraints. But, refer to~\citep{huang2023contrastive}, the $\mathcal{L}_{{1}}$ loss may lead the student model to overfit on incorrect predictions, thereby resulting in confirmation bias. To address this issue, ~\citep{huang2023contrastive} not only introduced a reliable bank to store pseudo-labels, which are determined by the NR-IQA method to select the ``best-ever'' quality outputs from the teacher network as reliable pseudo-labels, but also replaced the typical $\mathcal{L}_{1}$ loss used in consistency regularization with a contrastive loss.
Nevertheless, only the NR-IQA metric is utilized to determine ``best-ever'' outputs, focusing primarily on visual quality while neglecting corresponding image content, which may be inadequate for conditional diffusion related to image content.
% -------------------------------------------------------------sample of WaveDiff---------------------------------------------------------------------------
\begin{figure*}[htbp!]
\begin{center}
\includegraphics[width=1.0\linewidth]{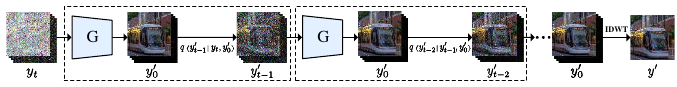}
\end{center}
\caption{The denoising process of WaveDiff. The operations within the dashed box will be iteratively repeated until obtaining the final denoised output $q(y_0')$}
\label{fig:sample}
\end{figure*}
% -------------------------------------------------------------sample of WaveDiff---------------------------------------------------------------------------
% -------------------------------------------------------------training process of WaveDiff-----------------------------------------------------------------
\begin{algorithm}[t!] 
  \caption{The training process of WaveDiff}
  \label{Alg:DDGAN} 
  \footnotesize 
  \setlength{\belowcaptionskip}{-3.5mm}
  \begin{algorithmic}[1] 
  \itemsep 0.1em % 数字表示行号间的间
  \Repeat
    \State $y \sim q(y), t \sim \mathrm{Uniform}(\{1, \dotsc, T\}), z \sim \mathcal{N}(0, \mathbf{I})$
    \State \textbf{Forward process $\to$ diffusion:}
    \State $y_0=\textbf{DWT}(y)$
    \State $q(y_{t-1} \mid y_0) = \mathcal{N}(y_{t-1}; \sqrt{\bar{\alpha}_{t-1}} y_0, 1 - \bar{\alpha}_{t-1})$
    \State $q(y_t \mid y_{t-1}) = \mathcal{N}\left(y_t; \sqrt{\alpha_{t}} y_{t-1}, 1 - \alpha_{t}\right)$
    \State \textbf{Reverse process $\to$ denoising:}
    \State $y_0' = G(y_t, z, t)$ 
    \State$q(y'_{t-1} \mid y_t, y_0') = \mathcal{N}\left(y'_{t-1}; \tilde{\mu}_t(y_t, y_0'), \tilde{\beta}_t\right)$
    \State \textbf{Model Optimization:}
    \State $ \text{Min}\mathcal{L}_{\text{adv}}^D = -\log D(y_{t-1}, y_t, t) + \log D(y'_{t-1}, y_t, t)$ 
    \State $\text{Min}\mathcal{L}_{\text{adv}}^G = -\log D(y'_{t-1}, y_t, t)+\mathcal{L}_{1}(y_0',y_0)$
    \Until{converged}
  \end{algorithmic}
\end{algorithm}
% \vspace{-4mm}
% -------------------------------------------------------------training process of WaveDiff-----------------------------------------------------------------
\subsection {Diffusion-based Generative Models}
Diffusion models, a novel class of generative models, have been recognized as SOTA generators due to their effective learning of complex distributions and their strong potential for generating diverse, high-fidelity samples. However, it has always suffered from low sampling efficiency. On this issue, Xiao \textit{et al.} \citep{xiao2021tackling} claimed that the slow sampling of diffusion models stems from the assumption that the denoising distribution is approximated by Gaussian distributions. For that, they introduce a novel generative model, termed DiffusionGAN, to parameterize the denoising distribution with a more expressive multimodal distribution, enabling denoising for larger steps. Although this model bridges the speed gap with GAN, it is still largely slower than GAN competitors. {Building upon this, Phung \textit{et al.}~\citep{phung2023wavelet} integrating the wavelet-based diffusion scheme into the diffusion process, namely WaveDiff, which offers faster convergence than the baseline Diffusion, and closing the gap with StyleGAN models~\citep{HUANG2023272,LEE2024106271}.} Unlike traditional diffusion models, WaveDiff introduces a novel wavelet-based generative model to parameterize the denoising distribution rather than assuming the denoising distribution is Gaussian distribution, thus enabling large step sizes for fast sampling, and the diffusing/denoising process is performed in the wavelet domain.

The training process of WaveDiff is outlined in algorithm \ref{Alg:DDGAN}, wherein the denoising process is illustrated in Fig.~\ref{fig:sample}. Specifically, the model comprises a generator and a discriminator, operating on images in the wavelet domain. Given the input $y$, it is first transformed into $y_0$ through wavelet transformation and then fed into the model for training. The generator is used to predict the denoised output of $y_{t-1}'$. But instead of directly outputting $y_{t-1}'$ in the denoising step, the generator first predicts the $y_0'$ from the noise-corrupted version of the data $y_t \sim q(y_t \mid y_0)$ at time step t, and then using the posterior distribution: $y_{t-1}' \sim q(y_{t-1}' \mid y_t, y_0')$ to get the denoised output $y_{t-1}'$.  So it can implicitly model the denoising distribution. The discriminator is used to determine the authenticity of the denoised outputs and optimizes both the generator and the discriminator in an adversarial training manner. Moreover, the introduction of wavelets enrich the frequency-aware at both image and feature levels, which will facilitate the perceptual quality of the generated images \citep{ozdenizci2023restoring}. Inspired by this model, we customize a wavelet diffusion model, built upon WaveDiff, to serve as the backbone of SemiDDM-Weather by modifying its inputs and loss functions to better adapt to our adverse weather removal task.

% -------------------------------------------------------------denoising process of WaveDiff----------------------------------------------------------------
\begin{figure*}[htbp!]
\begin{center}
\includegraphics[width=1.0\linewidth]{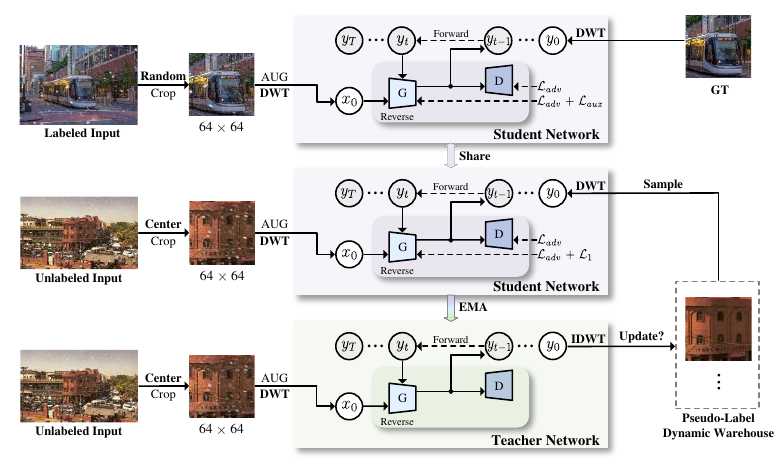}
\end{center}
\vspace{2mm}
\caption{Illustration of our \textbf{SemiDDM-Weather} framework, which innovatively integrates a DDM into the teacher-student network
}
\label{fig:total}
\end{figure*}

% -------------------------------------------------------------denoising process of WaveDiff----------------------------------------------------------------

\section{The Proposed SemiDDM-Weather} 
This section detailedly describes the design of our SemiDDM-Weather. Section {\ref{sec:tol}} outlines the proposed semi-supervised adverse weather removal framework, i.e., SemiDDM-Weather. {How we customize a wavelet diffusion model to serve as the backbone of SemiDDM-Weather}, is provided in Section \ref{sec:wavediff}. The targeted design for improving the accuracy of pseudo-labels under the teacher-student based network is detailed in Section \ref{sec:_labels}. Training and inference details are presented in Section \ref{sec: training and inference}.

\subsection{Semi-Supervised Adverse Weather Removal Framework}
\label{sec:tol}
To achieve the removal of various adverse weather conditions with limited labeled data, inspired by study in~\citep{sohn2020fixmatch}, {we adopt a teacher-student network\footnote{In this paper, for differentiation, we use the subscripts `$\text{tn}$' and `$\text{sn}$' to indicate the teacher and student networks, respectively.} as the foundation for semi-supervised learning}. {As depicted in Fig.~\ref{fig:total}, the teacher and student networks share the same structure network with a customized wavelet diffusion model as the backbone} to effectively capture the complex many-to-one mapping distributions. 
The teacher network is responsible for generating pseudo-labels for unlabeled images, which are subsequently utilized to guide the training of the student network. Its weights $\theta_{\text{tn}}$ are updated from the student network via EMA strategy, i.e.,
\begin{equation}
\theta_{\text{tn}} = \eta\cdot\theta_{\text{tn}} + (1 - \eta)\cdot\theta_{\text{sn}}, 
\end{equation}
%\vspace{-0.2em}
where $\eta \in (0, 1)$ denotes the momentum that controls the updating rate, serving to smooth the parameters over time. Given that the teacher network's outputs may not consistently outperform the student network's, directly using them as pseudo-labels could introduce inaccuracies and potentially mislead the student network's training. 
{To abate this risk, we use the “optimal” outputs generated by the teacher network during training to more effectively steer the student network's learning.}

The student network parameters, denoted as $\theta_{\text{sn}}$, are trained with both labeled images and unlabeled images. Concretely, the training of the student network can be formulated as minimizing the loss $\mathcal{L}$:
\begin{equation}
    \mathcal{L} =\mathcal{L}_{\text{sup}} + \lambda_t \cdot \mathcal{L}_{\text{unsup}},
\end{equation}
where $\mathcal{L}_{\text{sup}}$ and $\mathcal{L}_{\text{unsup}}$ represent the supervised loss and unsupervised loss, respectively, and $\lambda_t$ is a time-varying hyper-parameter for the trade-off between $\mathcal{L}_{\text{sup}}$ and $\mathcal{L}_{\text{unsup}}$.

\subsection{Customization of a Wavelet Diffusion Model} 
\label{sec:wavediff}
In our scheme, degraded images are required as conditional inputs to the generator, allowing us to generate pairwise corresponding estimates of these degraded images, but this inevitably consumes more computational resources. Moreover, {WaveDiff's generator exclusively utilizes $\mathcal{L}_{1}$ loss for constraining the generated content}, which does not adequately ensure content consistency between the generated outputs and the degraded image conditional inputs. {Therefore, directly using it as the backbone of our framework may not be well-suited to our task. Upon these deficiencies, We customize a wavelet diffusion model as the backbone of SemiDDM-Weather which is built upon WaveDiff by modifying its model inputs and loss functions to better adapt to our adverse weather removal task.} Specifically, to trade off the computational resource consumption and high-performance model training, we adopt a crop-based training strategy that feeds the model with a cropped $64 \times 64$ patch from the input image instead of the entire image. To address the limitations of using only $\mathcal{L}_{{1}}$ loss for content consistency constraint, we incorporate perceptual loss to guide training, which can effectively preserve the texture and structural information of the images. In addition, we further extend the $\mathcal{L}_{{1}}$ loss to a combination of $\mathcal{L}_{{1}}$ and $\mathcal{L}_{{2}}$ losses, thus combining the advantages of the $\mathcal{L}_{{1}}$ loss in maintaining pixel accuracy with the insensitivity of the $\mathcal{L}_{{2}}$ loss to noise for better content reconstruction. What's more, similar to previous semi-supervised learning frameworks, we use the contrast loss instead of the $\mathcal{L}_{{1}}$ loss as the consistency constraint to reduce the risk of the model overfit to incorrect predictions, thereby diminishing confirmation bias~\citep{huang2023contrastive}.
%-----------------------------------------------------------acquire pseudo-labels---------------------------------------------------------------------------
\begin{algorithm}[t] 
  \caption{The acquisition of ``optimal'' pseudo-labels}
  \label{Alg:bank} % 算法的标题
  \footnotesize % 字体大小调整为小号
  \begin{algorithmic}[1] 
  \itemsep0.4mm % 数字表示行号间的间隔
  % \State \textbf{Update reliable bank:}
     \State ${y}^t_c=\text{Reverse}_t(y_t,x_0),{y}^s_c=\text{Reverse}_s(y_t,x_0)$
   \If{$\Call{Q}{{y}^t_c} > \max(\Call{Q}{{y}^s_c}, \Call{Q}{y_c}) \text{ and } \mathcal{L}_1({y}^t_c, {y}^s_c) < \varphi$}
    \State ${y}^r_c = {y}^t_c$ 
\Else
    \State ${y}^r_c = y_c$
\EndIf
 \end{algorithmic}
\end{algorithm}
% \vspace{-4mm}

%-----------------------------------------------------------acquire pseudo-labels---------------------------------------------------------------------------
\subsection{Target Design for Enhancing Pseudo-label Accuracy}
\label{sec:_labels}
{Since the teacher network’s outputs do not always surpass those of the student network, using its {current} outputs as pseudo-labels can potentially mislead the training of the student network on unlabeled data. To address this issue, the targeted design has been made to improve the accuracy of the pseudo-labels used to guide the student network's training.
Specifically, we utilize the ``optimal'' outputs obtained from the teacher network during the training process as the pseudo-labels to guide the student network training with unlabeled images, rather than relying on the current outputs. All of these ``optimal'' outputs are stored in the pseudo-label dynamic warehouse.
Regarding the construction of this dynamic warehouse, we first employ the model, trained exclusively on labeled data, to establish the initial ``optimal'' output for each piece of unlabeled data. Next, in each training iteration, we compare the current output of the teacher network with both the output of the student network and the corresponding ``optimal'' one in the pseudo-label dynamic warehouse. If the teacher's current output is of the best quality, we will immediately use it to update the ``optimal'' pseudo-label stored in the dynamic warehouse. This process ensures that the outputs stored in the pseudo-label dynamic warehouse consistently represent the optimal results produced by the teacher network throughout training.}

As to evaluating whether the teacher network's current output is of the best quality, we employ two criteria. 1) An NR-IQA metric, MUSIQ~\citep{ke2021musiq}, is introduced to evaluate the visual quality. The quality of the teacher network's current output is proportional to the score of MUSIQ, i.e., the better the quality, the higher the score. As shown in line 2 of Algorithm~\ref{Alg:bank},~$\Call{Q}{{x}}$ represents the MUSIQ score of input $x$. 2) A consistency constraint is incorporated to evaluate the quality of teacher's outputs. As shown in line 2 of Algorithm~\ref{Alg:bank}, the $\mathcal{L}_{{1}}$ distance between the outputs of the teacher and the student networks is required to be less than a predefined threshold $\varphi$. Based on extensive testing, $\varphi$ is optimally set to 0.1 in our scheme. Note that the predictions of the teacher network were considered to be of the best quality only if they comply with these two criteria.

%-----------------------------------------------------------training process of Semi------------------------------------------------------------------------
\begin{algorithm}[t!] 
  \caption{The training process of \textbf{SemiDDM-Weather}}
  \label{Alg:supervised} % 算法的标题
  \footnotesize % 字体大小调整为小号
  \begin{algorithmic}[1] 
  \itemsep0.1em % 数字表示行号间的间隔
   \Require
    \State degraded image $x$, label $y$ (labeled data: GT $\to$ $y$; unlabeled data: ``optimal'' pseudo-label $\to$ $y$)
    \Repeat
    \State $x_{c}={\text{Crop}}_{64}({x}),x_0=\text{DWT}(x_{c})$
    \State $y_{c}={\text{Crop}}_{64}({y}),y_0=\text{DWT}(y_{c})$
    \State $y_0 \sim q(y_0), t \sim \mathrm{Uniform}(\{1, \dotsc, T\}), z \sim \mathcal{N}(0, \mathbf{I})$
    \State \textbf{Forward process $\to$ diffusion:}
    \State $q(y_{t-1} \mid y_0) = \mathcal{N}(y_{t-1}; \sqrt{\bar{\alpha}_{t-1}} y_0, 1 - \bar{\alpha}_{t-1})$
    \State $q(y_t \mid y_{t-1}) = \mathcal{N}\left(y_t; \sqrt{\alpha_{t}} y_{t-1}, 1 - \alpha_{t}\right)$
    \State \textbf{Reverse process $\to$ denoising:}
    \State ${y}_0' = G_{\text{sn}}(y_t, x_0, z, t)$
    \State$q({y}_{t-1}' \mid y_t, {y}_0') = \mathcal{N}\left({y}_{t-1}'; \tilde{\mu}_t(y_t, {y}_0'), \tilde{\beta}_t\right)$
    \State ${y}_{c}'=\text{IDWT}( {y}_0')$
   \If {{Labeled}}
   \State$\mathcal{L}_{\text{aux}}=\mathcal{L}_{\text{rec}}({y}_c',y_c)+\mathcal{L}_{\text{perc}}({y}_c',y_c)$
   \State $\mathcal{L}^G_{\text{sup}} = -\log D_{\text{sn}}({y}_{t-1}', y_t, t)+\mathcal{L}_{\text{aux}}({y}_c',y_c)$ 
   \State $\mathcal{L}^D_{\text{sup}} = -\log D_{\text{sn}}(y_{t-1}, y_t, t) + \log D_{\text{sn}}({y}_{t-1}', y_t, t)$
   \State $\text{Minimize} ~\mathcal{L}^G_{\text{sup}}, ~\text{Minimize} ~\mathcal{L}^D_{\text{sup}}$
   \Else
   \State ${y}^t_c=\text{Reverse}_t(y_t,x_0),{y}^s_c=\text{Reverse}_s(y_t,x_0)$
   \State $y^r_c=\text{Optimal}(y_c,{y}^t_c,{y}^s_c)$
   \State $\mathcal{L}^G= -\log D_{\text{sn}}({y}_{t-1}', y_t, t) + \mathcal{L}_{1}({y}_0',y_0)$
   \State $\mathcal{L}^G_{\text{sup}} = \mathcal{L}^G + \mathcal{L}_{\text{contr}}({y}^s_c,{y^r_c},x_c)$   
   \State $\mathcal{L}^D_{\text{sup}} = -\log D_{\text{sn}}(y_{t-1}, y_t, t) + \log D_{\text{sn}}({y}_{t-1}', y_t, t)$
   \State $\text{Minimize}~\lambda_t \cdot \mathcal{L}^G_{\text{sup}}, ~\text{Minimize}~\lambda_t \cdot \mathcal{L}^D_{\text{sup}}$
    \EndIf
    \Until{converged}
  \end{algorithmic}
\end{algorithm}
%-----------------------------------------------------------training process of Semi------------------------------------------------------------------------

\subsection{Training and Inference details} 
\label{sec: training and inference}
\begin{figure}[t!]
\centering
\includegraphics[width=0.9\linewidth]{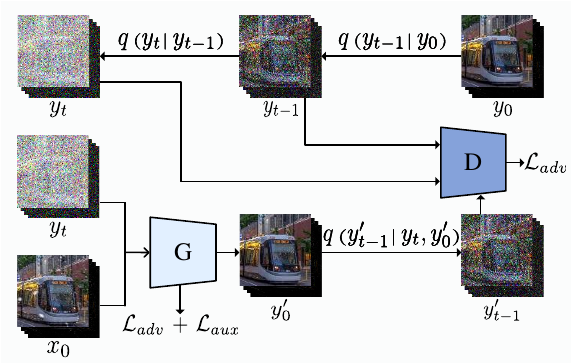}
\caption{The schematic diagram of labeled data training in the student network}
\label{fig:labeled}
\end{figure}

In the following, we will describe the model training and inference in detail. The training process is outlined in Algorithm~\ref{Alg:supervised}. The schematic diagram of labeled data training is shown in Fig.~\ref{fig:labeled}.
To be specific, given the degraded image \( x \in \mathbb{R}^{3 \times \text{H} \times \text{W}} \) and its corresponding label \( y \in \mathbb{R}^{3 \times \text{H} \times \text{W}} \), where \( y \) represents the GT if \( x \) is labeled data, otherwise the {``optimal'' outputs from the pseudo-label dynamic warehouse}, we first crop them into $64 \times 64$ patches, resulting  \( x^{c} \in \mathbb{R}^{3 \times 64 \times 64} \) and \( y^{c} \in \mathbb{R}^{3 \times 64 \times 64} \). Note that the labeled data is randomly cropped while the unlabeled ones are center-cropped, due to the consideration of the alignment of the ``optimal'' outputs. Next, we decompose them into four wavelet subbands and then concatenate them into a single target in channel dimension, i.e., \( x_0 \in \mathbb{R}^{12 \times 64 \times 64} \) and \( y_0 \in \mathbb{R}^{12 \times 64 \times 64} \), for the diffusion/denoising process. More concretely, referring to Fig.~\ref{fig:labeled}, we inject Gaussian noise to \(y_{0}\) (i.e., diffusion), obtaining \(y_{t-1}\) and \(y_{t}\), as detailed in lines 7-8 of Algorithm~\ref{Alg:supervised}.
Subsequently, the noise-corrupted result $y_{t}$ and the degraded image \(x_{0}\) are fed into the generator, resulting in the predicted \({y}_0'\). This predicted value is then used in the posterior distribution: \({y}_{t-1}' \sim q({y}_{t-1}' \mid y_t, {y}_0')\), for denoising, yielding the denoised output \({y}_{t-1}'\), as shown in lines 10-11 of Algorithm~\ref{Alg:supervised}. 
% -------------------------------------------------------------training process of labeled data-------------------------------------------------------------
% -------------------------------------------------------------training process of labeled data-------------------------------------------------------------
For labeled image training, considering that we are utilizing a conditional diffusion model to address various adverse weather degradation issues, it is important to maintain the content consistency between the generated outputs and the degraded image conditional inputs.
To achieve this, we refine the loss function of the WaveDiff's generator, i.e., 
\begin{equation}\label{eq:content}
\mathcal{L}^G_{\text{sup}} = -\log D_{\text{sn}}({y}_{t-1}', y_t, t) + \mathcal{L}_{\text{aux}}({y}_c',y_c).
\end{equation}
Referring to Eq. \eqref{eq:content}, apart from the original adversarial loss $\log D_{\text{sn}}({y}_{t-1}', y_t, t)$ in WaveDiff, a auxiliary loss $\mathcal{L}_{\text{aux}}({y}_c',y_c)$ comprises of $\mathcal{L}_{\text{rec}}$ loss and $\mathcal{L}_{\text{perc}}$ are further introduced. Specifically, the $\mathcal{L}_{\text{aux}}$ is formulated as:
\begin{equation}
    \mathcal{L}_{\text{aux}} =5\cdot\mathcal{L}_{\text{rec}}({y}_c',y_c) + 10\cdot\mathcal{L}_{\text{perc}}({y}_c',y_c),
\end{equation}
where
\begin{equation}
    \mathcal{L}_{\text{rec}} =0.1\cdot\mathcal{L}_{1}({y}_c',y_c) + 0.9\cdot\mathcal{L}_{2}({y}_c',y_c),
\end{equation}
% \paragraph{$\mathcal{L}_{perc}$ Loss}
\begin{equation}
     \mathcal{L}_{\text{perc}}= \sum_{j=1}^{K} \sum_{i=1}^{N} | \phi_j({y'_c}_i) - \phi_j({y_c}_i)|.
\end{equation}
$\mathcal{L}_{\text{rec}}$ represents the reconstruction loss, it a combination of $\mathcal{L}_{1}$ loss and $\mathcal{L}_{2}$ loss. $\mathcal{L}_{\text{perc}}$ denotes the perpetual loss, obtained by using a pre-trained VGG-16 network~\citep{simonyan2014very}. $\phi_j$ is the $j^{th}$ pooling layer's activation map in the VGG-16, and this module selects pool-1, pool-2, and pool-3. $y_c$ is the crop patch of the labeled image and $y'_c$ is the corresponding prediction of the generator. $D_{\text{sn}}({y}_{t-1}', y_t, t)$ is the original adversarial loss of WaveDiff, defined as:
\begin{equation}
D_{\text{sn}}({y}_{t-1}', y_t, t) = \sigma({\mathcal{FC}}({\mathcal{OP}}({cat}(y_{t-1}',y_t), t))),
\end{equation}
where ${cat} (y_{t-1}', y_t)$ denotes the operation of concatenating $y_{t-1}'$ and $y_t$ along a specified dimension. The function ${\mathcal{OP}}$ refers to a series of operations, typically involving convolutional layers followed by batch normalization and non-linear activations, which are designed to extract and refine features from the concatenated tensor. The term $\mathcal{FC}$ denotes the final fully connected layer, which receives these processed features and outputs a scalar value utilized for classification purposes. The function $\sigma$ denotes the sigmoid activation, which maps the output of the $\mathcal{FC}$ layer to a probability between 0 and 1.

As for the discriminator loss, as shown in line 16 of Algorithm~\ref{Alg:supervised}, it follows exactly the one of WaveDiff. The specific form of the discriminator loss is as follows:
\begin{equation} \label{D_loss}
\mathcal{L}^D_{\text{sup}} = -\log D_{\text{sn}}(y_{t-1}, y_t, t) + \log D_{\text{sn}}({y}_{t-1}', y_t, t).
\end{equation}

% --------------------------------------------------------------------crop strategy-------------------------------------------------------------------------
\begin{figure}[t!]
\centering
\includegraphics[width=1.0\linewidth]{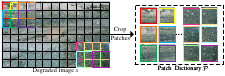}
\caption{The schematic diagram of the dense regular grid cropping strategy. In this diagram, the grid size represents the spacing used for cropping, while the box size represents the size of the cropped images. In our method, we set the spacing to 4 pixels, and each crop is sized at $64\times64$ pixels. The cropping sequence systematically proceeds along the grid from left to right and from top to bottom, and the cropped patches of $x$ are collected in a dictionary $\mathcal{P} = \{{x}_{p_1}, {x}_{p_2}, ..., {x}_{p_K}\}, {x}_{p_k}={\text{Crop}}_{64}(\mathbf{M}_{k} \circ x)$, where $\circ$ indicates element-wise multiple and $\mathbf{M}_{k}$ is the $k^{th}$ patch mask of $x$ (1 for corresponding patch locations while 0 for others)
} 
\label{fig:crop}
\end{figure}
%-------------------------------------------------------------------model_inference-------------------------------------------------------------------------
For unlabeled data training, the discriminator loss, as shown in line 23 of
Algorithm 2, also follows the formulation used in WaveDiff, i.e., Eq. \eqref{D_loss}.
For the generator loss, as shown in line 21 of Algorithm~\ref{Alg:supervised}, only the $\mathcal{L}_{\text{1}}$ loss and the adversarial loss are employed to train the generator, as it is considered that applying too many constraints to unlabeled data with pseudo-labels may mislead the training process. 
Besides, as aforementioned, simply using the $\mathcal{L}_{{1}}$ distance as the consistency loss can easily overfit the student model to incorrect predictions, thus resulting in confirmation bias. To cope with this issue, we further replace it with a contrastive loss $\mathcal{L}_{\text{contr}}$ in our scheme, i.e.,
\begin{equation} \label{eq:contr}
   \mathcal{L}_{\text{contr}} = \sum_{j=1}^{K} \sum_{i=1}^{M} \omega_j \cdot\frac{|\phi_j({y^s_c}(I)), \phi_j({y^r_c}(i))|}{|\phi_j({y^s_c}(i)), \phi_j({x_c}(i))|},
\end{equation}
where $x_c$ represents the unlabeled degraded image, ${y^s_c}$ is the denoising output from the student network and ${y^r_c}$ is the corresponding ``optimal'' output. The function $\phi_j$ denotes the $j^{th}$ hidden layer of the pre-trained VGG-19 network, and $\omega_j$ is the corresponding weight coefficient. The $\mathcal{L}_{{1}}$ distance is employed in Eq. \eqref{eq:contr} to quantify the feature space disparity between the students’ denoising outputs with a positive sample (``optimal'' output) and a negative sample (degraded input image).
%-------------------------------------------------------------------model_inference-------------------------------------------------------------------------
\begin{figure*}[t!]
\centering
\includegraphics[width=1.0\linewidth]{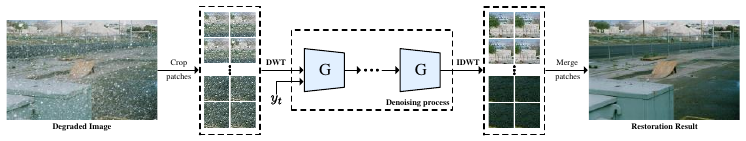}
\caption{The schematic diagram of the model inference process. For the denoising process, please refer to Fig.~\ref{fig:sample} for a similar process. The cropping process can similarly reference Fig.~\ref{fig:crop}}
\label{fig:denoising}
\end{figure*}

%-------------------------------------------------------------------model_inference-------------------------------------------------------------------------
%-------------------------------------------------------------------model_inference-------------------------------------------------------------------------
\begin{algorithm}[t!] 
  \caption{The inference process of \textbf{SemiDDM-Weather}}
  \label{Alg:inference} % 算法的标题
  \footnotesize % 字体大小调整为小号
  \begin{algorithmic}[1] 
  \itemsep0.1em % 数字表示行号间的间隔
   \renewcommand{\algorithmicrequire}{\textbf{Input:}}
   \Require degraded image $x$,  {patch mask dictionary $\mathbb{M}$}.
    \State  $t \sim \mathrm{Uniform}(\{1, \dotsc, T\}), z \sim \mathcal{N}(0, \mathbf{I})$
    \State $\bm{\hat{\Omega}}_t=\mathbf{0}, \mathbf{M}=\mathbf{0}$
    \For {$k = 1,\ldots,K$}
        \State $x_{p_k}={\text{Crop}}_{64}(\mathbf{M}_k \circ x$), $\mathbf{M}_k \in \mathbb{M}$ // $\circ$: element-wise multiplication
        \State $x_0=\text{DWT}(x_{p_k})$, 
        \State $y_t \sim \mathcal{N}(0, \mathbf{I})$, 
        \For {$t = T,\ldots,1$}
            \State $y_0' = G_{\text{sn}}(y_t, x_0, z, t)$
            \State $q(y_{t-1}' \mid y_t, y_0')$
        \EndFor
        \State $y_{p_k}'=\text{IDWT}(y_0')$
        \State {$\bm{\hat{\Omega}}_t = \bm{\hat{\Omega}}_t + Filling(\mathbf{M}_k, y_{p_k}'$})  // fill the patch in $\mathbf{M}_k$ with $y_{p_k}'$
        \State $\mathbf{M} = \mathbf{M} + \mathbf{M}_k$
    \EndFor
    \State $\bm{\hat{\Omega}}_t = \bm{\hat{\Omega}}_t \oslash \mathbf{M}$ \quad  \quad \quad  \quad \quad \quad// $\oslash$: element-wise division
    \State \Return $\bm{\hat{\Omega}}_t$
  \end{algorithmic}
\end{algorithm}
% \vspace{-2mm}
%-------------------------------------------------------------------model_inference-------------------------------------------------------------------------
For model inference, we continue to use the crop-based data input strategy. Notably, there is a slight difference. Instead of randomly cropping, we crop the degraded image $x$ into $K$ overlapping patches of size 64$\times$64 over dense regular grids with a spacing of 4 pixels, {collected into a dictionary $\mathcal{P}$}. The dense regular grid cropping strategy is illustrated in Fig.~\ref{fig:crop}. These patches will be firstly transformed by wavelet and then input into the student network for denoising. After that, the inverse wavelet transform will be conducted on the denoised results to obtain the restoration of these patches. Finally, the restored patches will be reassembled to reconstruct the entire image. Note that, for the overlapping regions, an averaging method is applied to ensure visual consistency, effectively smoothing the edges of these areas to achieve a cohesive and natural image restoration outcome. The illustration of this process is referred to in Fig.~\ref{fig:denoising}.
As for the details of denoising process, we fed the variable $y_t$, the conditional input $x_0$, the latent variable $z \sim \mathcal{N}(0, \mathbf{I})$, and the timestep $t$ into the generator of the student network to approximate the $y_0' = G_{\text{sn}}(y_t, x_0, z, t)$. According to the posterior distribution $q(y_{t-1}' \mid y_t, y_0')$, then we can obtain the denoised output $q(y_{t-1}')$. This process is iteratively repeated until obtaining the final denoised output $q(y_0')$, as shown in Algorithm \ref{Alg:inference}, steps 7-10.

\section{Experimental Settings}
(1) \textit{Data Preparation}: In this part, we detailedly describe the datasets with challenging adverse weather conditions used in our method. Similar to other all-in-one methods~\citep{ozdenizci2023restoring,valanarasu2022transweather,zhu2023learning}, we use a combination of several public adverse weather datasets for training, including Raindrop, Outdoor-Rain, and Snow100K.

{Training Dataset.}~~\textit{Raindrop}~\citep{qian2018attentive} training set, comprising 861 images with raindrops, introduce real raindrop artifacts on the camera sensor to obstruct the view;
\textit{Outdoor-Rain}~\citep{li2019heavy} training set, with 9000 synthetic images impaired by both fog and rain streaks; and \textit{Snow100K}~\citep{liu2018desnownet} training set, compromise 50000 synthetic images degraded by snow. Given the number of images in each dataset is not equal, for the Snow100k dataset, only 9000 images are randomly selected for training. Note that for each dataset, we divided it into labeled and unlabeled sets at a ratio of 1:1.

{Test Dataset.}~~\textit{Raindrop}~\citep{qian2018attentive} test set, containing 58 images, was acquired in the same manner as the training set; 
\textit{Outdoor-Rain}~\citep{li2019heavy} test set, including 750 synthetic images;
\textit{Snow100K-2000}~\citep{liu2018desnownet} test set, selecting 2000 synthetic images from \textit{Snow100K}'s test set like~\citep{cui2023focal,cui2023selective}. {\textit{Snow100K-Real}~\citep{liu2018desnownet} dataset, comprising 1329 authentic snow-degraded test images}; and \textit{IVIPC-DQA}~\citep{wu2020subjective} dataset, involving 206 real-scene images under rain conditions. 

(2) \textit{Implementation Details}: Our method is implemented on the Pytorch platform and conducted on NVIDIA RTX 3090 GPUs. AdamP~\citep{heo2020adamp} is adopted as our optimizer. The model training consists of two phases, wherein the first phase lasts 500 epochs for a batch size of 196, and the second one extends to 650 epochs for a batch size of 64. Specifically, in the first phase, we train the model only by using labeled data and then use the trained model to initialize the pseudo-label dynamic warehouse to get the initial ``optimal'' pseudo-labels for unlabeled data. In the second phase, we restart the training by incorporating unlabeled data and its corresponding ``optimal'' pseudo-labels. In both phases, the initial learning rate is set to {$1.6\mathrm{e}^{-4}$} and $1.25\mathrm{e}^{-4}$ for the generator and the discriminator, respectively. 
Considering that our model is trained on a composite dataset, which contains various weather-conditioned datasets, to ensure our method is robust enough to restore various types of weather-degraded images in the all-in-one model, we introduce the Memory Replay Training strategy \citep{kulkarni2022unified} for model training. To be specific, in the initial 150 training epochs, our method is exclusively trained using the \textit{Raindrop} dataset.  Upon completion of this phase, we extend the dataset collection to include the \textit{Outdoor-Rain} dataset and continue training for an additional 150 epochs. Finally, we enrich the training dataset by adding the \textit{Snow100K} dataset, which, along with the previously used datasets, is employed in the subsequent training. This strategy can make the model learn all types of weather restoration and give peak performance for each task~\citep{kulkarni2022unified}.   
\subsection{Comparison with SOTA Competitors}
We perform comparisons of our SemiDDM-Weather with several other fully-supervised SOTA methods, including weather-specific methods and all-in-one methods on the involved test datasets.
For weather-specific removal methods, we choose RaindropAttn~\citep{quan2019deep}, IDT~\citep{xiao2022image}, and UDR-S2Former~\citep{chen2023sparse} for raindrop removal, HRGAN~\citep{li2019heavy}, PCNet~\citep{jiang2021rain}, and MPRNet~\citep{zamir2021multi} for deraining, and FocalNet~\citep{cui2023focal}, SFNet~\citep{cui2023selective}, and FSNet~\citep{10310164} for desnowing, 
For all-in-one methods, we select TransWeather~\citep{valanarasu2022transweather}, WGWS-Net~\citep{zhu2023learning}, and WeatherDiff\textsubscript{64}~\citep{ozdenizci2023restoring}. 
The following detailed subjective and objective evaluation results demonstrate that despite using limited labeled data compared to these SOTA fully-supervised methods, our approach still achieves better adverse weather restoration results, even superior to the GT.

\begin{figure*}[htbp!]
\centering
\captionsetup[subfloat]{labelformat=empty} 
\subfloat[\scriptsize Input]{%
  \includegraphics[width=0.32\textwidth]{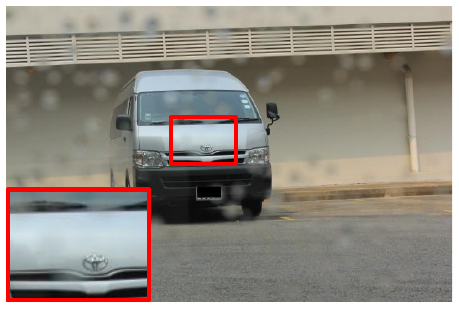}
}\hfill
\subfloat[\scriptsize RaindropAttn~\citep{quan2019deep}]{%
  \includegraphics[width=0.32\textwidth]{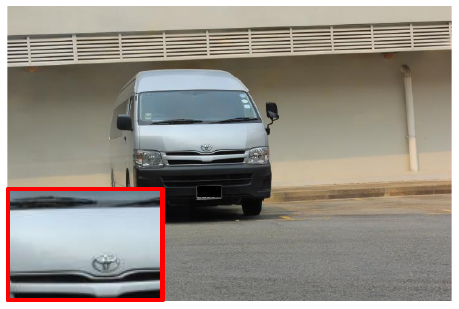}
}
\hfill
\subfloat[\scriptsize IDT~\citep{xiao2022image}]{%
  \includegraphics[width=0.32\textwidth]{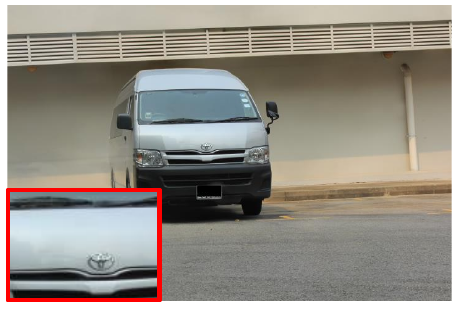}
}

\subfloat[\scriptsize UDR-S2Former~\citep{chen2023sparse}]{%
  \includegraphics[width=0.32\textwidth]{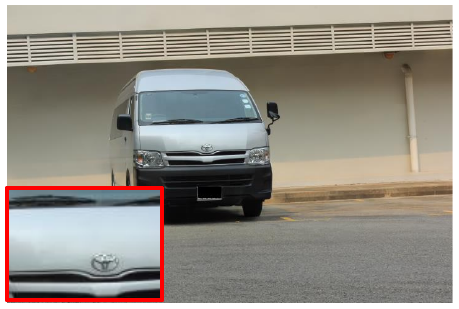}
}
\hfill
\subfloat[\scriptsize TransWeather~\citep{valanarasu2022transweather}]{%
  \includegraphics[width=0.32\textwidth]{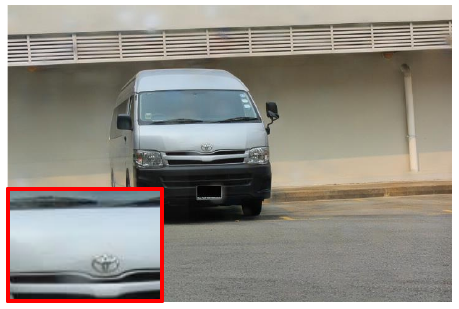}
}
\hfill
\subfloat[\scriptsize WGWS-Net~\citep{zhu2023learning}]{%
  \includegraphics[width=0.32\textwidth]{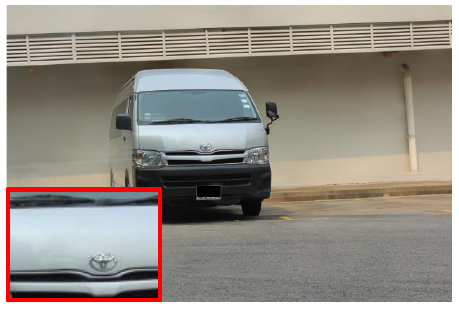}
}

\subfloat[\scriptsize WeatherDiff\textsubscript{64}~\citep{ozdenizci2023restoring}]{%
  \includegraphics[width=0.32\textwidth]{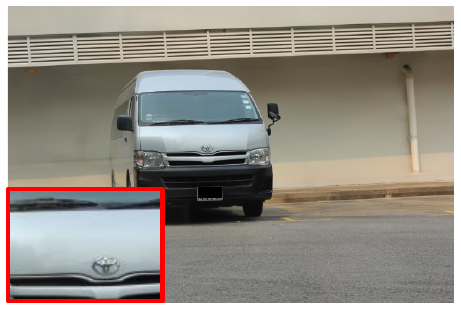}
}
\hfill
\subfloat[\scriptsize GT]{%
  \includegraphics[width=0.32\textwidth]{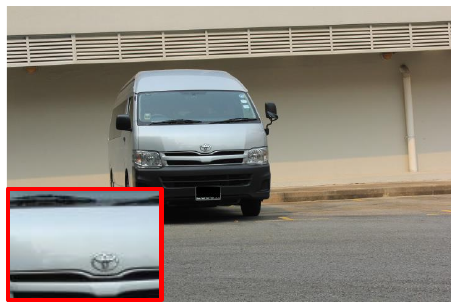}
}
\hfill
\subfloat[\scriptsize Ours]{%
  \includegraphics[width=0.32\textwidth]{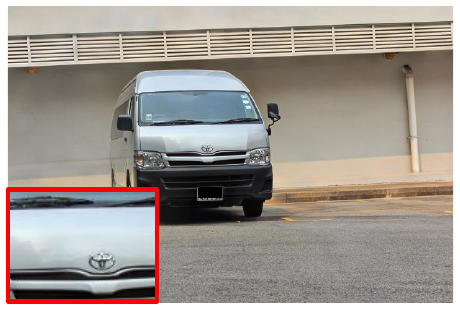}
}
\hfill
\caption{Restoration instances on~~\textit{Raindrop} test set by our method and other SOTA fully-supervised competitors. The red box highlights the zoomed-in patch}
% \vspace{-2mm}
\label{fig:compare_raindrop}
\end{figure*}

\begin{figure*}[htbp!]
\centering
\captionsetup[subfloat]{labelformat=empty}  % 这将移除子图的标签编号
% \captionsetup[subfloat]{captionskip=-4pt}
\subfloat[\scriptsize Input]{%
  \includegraphics[width=0.32\textwidth]{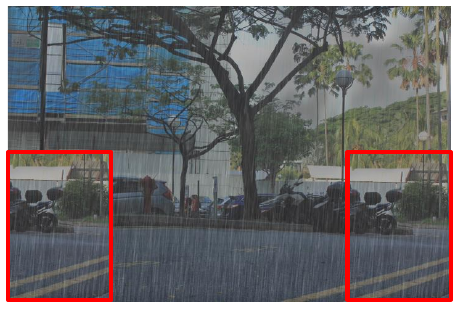}
  \label{fig:input3}
}\hfill
\subfloat[\scriptsize HRGAN~\citep{li2019heavy}]{%
  \includegraphics[width=0.32\textwidth]{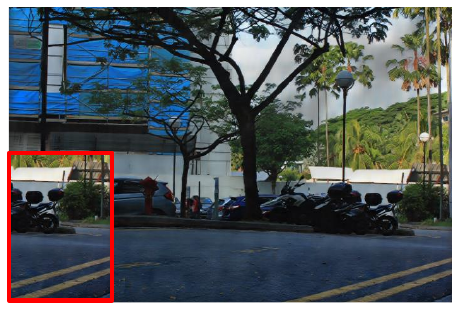}
}\hfill
\subfloat[\scriptsize PCNet~\citep{jiang2021rain}]{%
  \includegraphics[width=0.32\textwidth]{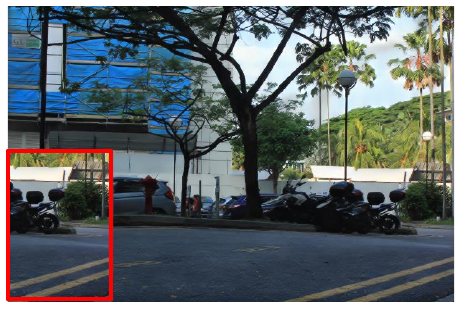}
}\hfill
\subfloat[\scriptsize MPRNet~\citep{zamir2021multi}]{%
  \includegraphics[width=0.32\textwidth]{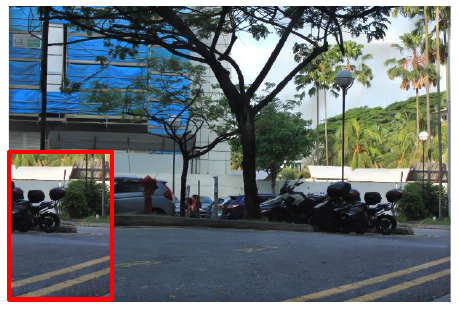}
}
\hfill
\subfloat[\scriptsize TransWeather~\citep{valanarasu2022transweather}]{%
  \includegraphics[width=0.32\textwidth]{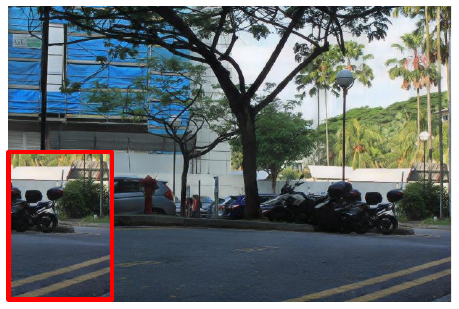}
}
\hfill
\subfloat[\scriptsize WGWS-Net~\citep{zhu2023learning}]{%
  \includegraphics[width=0.32\textwidth]{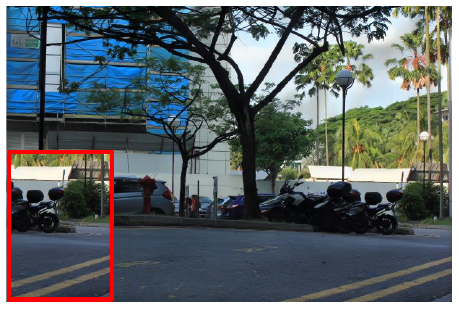}
}\hfill
\subfloat[\scriptsize WeatherDiff\textsubscript{64}~\citep{ozdenizci2023restoring}]{%
  \includegraphics[width=0.32\textwidth]{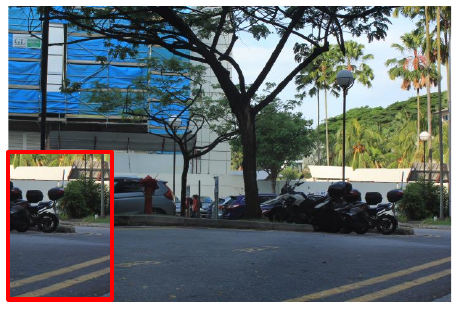}
}
\hfill
\subfloat[\scriptsize GT]{%
  \includegraphics[width=0.32\textwidth]{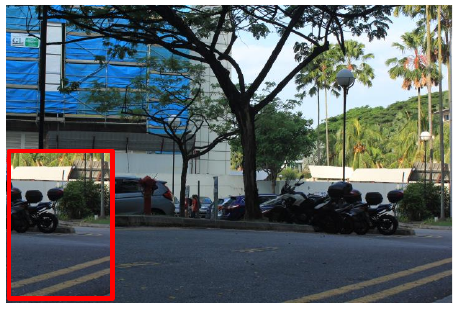}
}
\hfill
\subfloat[\scriptsize Ours]{%
  \includegraphics[width=0.32\textwidth]{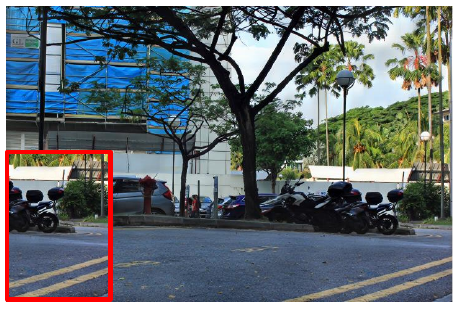}
}
\hfill
\caption{Restoration instances on~~\textit{Outdoor-Rain} test set by our method and other SOTA fully-supervised competitors. The red box highlights the zoomed-in patch}
\label{fig:compare_rain}
\end{figure*}

\begin{figure*}[htbp]
\centering
\captionsetup[subfloat]{labelformat=empty}  % 这将移除子图的标签编号
\subfloat[\scriptsize Input]{%
  \includegraphics[width=0.32\textwidth]{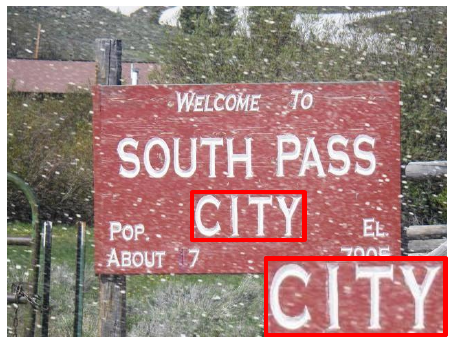}
  \label{fig:input5}
}
\hfill
\subfloat[\scriptsize FocalNet~\citep{cui2023focal}]{%
  \includegraphics[width=0.32\textwidth]{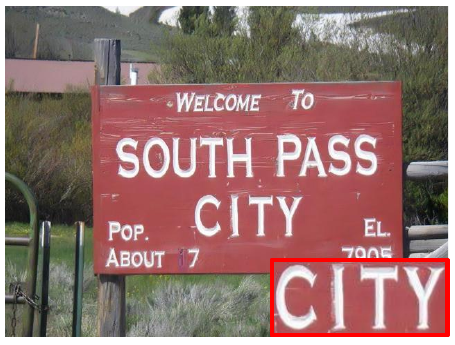}
}
\hfill
\subfloat[\scriptsize SFNet~\citep{cui2023selective}]{%
  \includegraphics[width=0.32\textwidth]{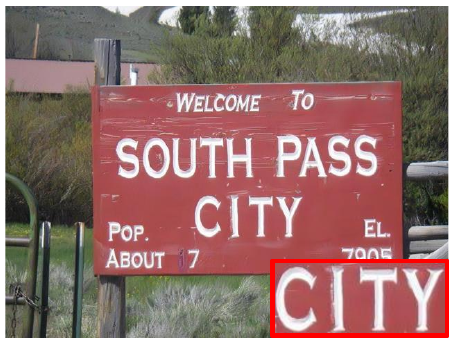}
}

\subfloat[\scriptsize FSNet~\citep{10310164}]{%
  \includegraphics[width=0.32\textwidth]{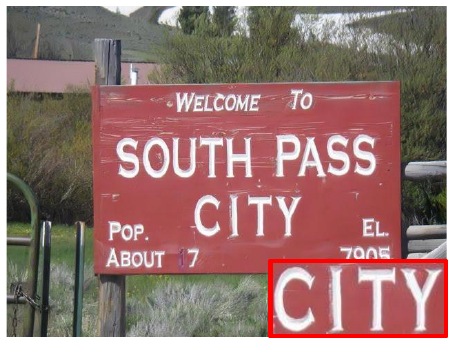}
}
\hfill
\subfloat[\scriptsize TransWeather~\citep{valanarasu2022transweather}]{%
  \includegraphics[width=0.32\textwidth]{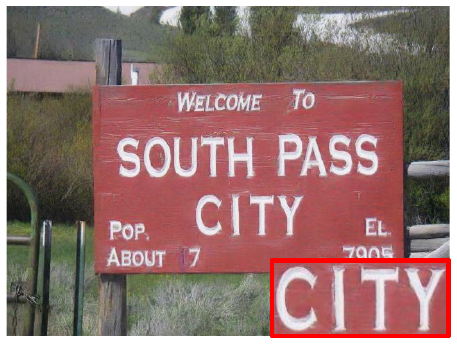}
}
\hfill
\subfloat[\scriptsize WGWS-Net~\citep{zhu2023learning}]{%
  \includegraphics[width=0.32\textwidth]{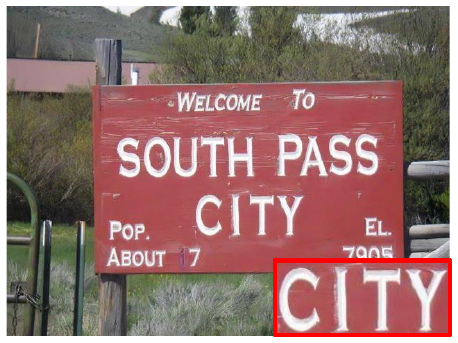}
}

\subfloat[\scriptsize WeatherDiff\textsubscript{64}~\citep{ozdenizci2023restoring}]{%
  \includegraphics[width=0.32\textwidth]{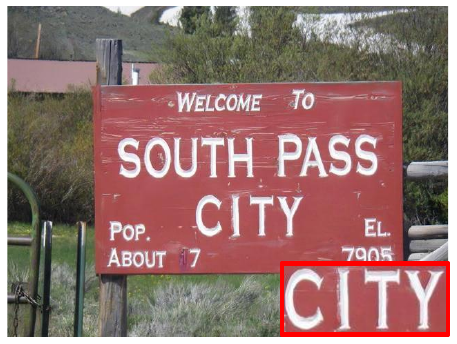}
}
\hfill
\subfloat[\scriptsize GT]{%
  \includegraphics[width=0.32\textwidth]{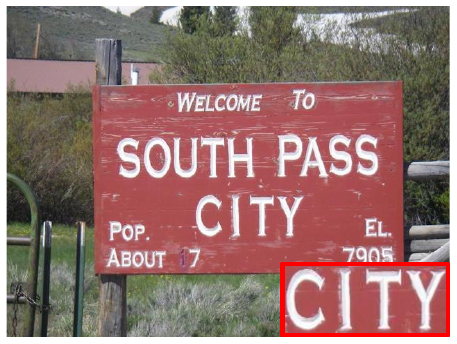}
}
\hfill
\subfloat[\scriptsize Ours]{%
  \includegraphics[width=0.32\textwidth]{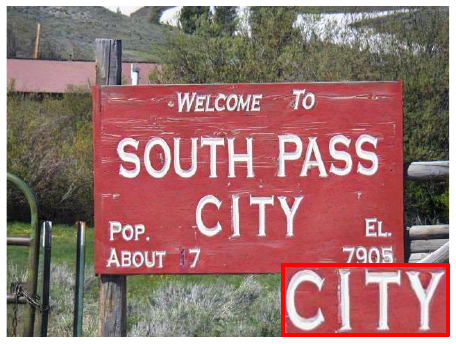}
}
\hfill
\caption{Restoration instances on~~\textit{Snow100K-2000} test set by our method and other SOTA fully-supervised competitors. The red box highlights the zoomed-in patch}
\label{fig:compare_snow}
\end{figure*}

(1) \textit{Subjective Evaluation}: Figs. \ref{fig:compare_raindrop}, \ref{fig:compare_rain}, and \ref{fig:compare_snow} illustrate instances of test datasets, including \textit{Raindrop}, \textit{Outdoor-rain}, and \textit{Snow100K-2000}. It shows that our method not only effectively removes the degradations caused by adverse weather conditions such as raindrops, rain streaks, and snow through a unified (a.k.a., all-in-one) framework, but also enhances the image clarity, such as clearer vehicle markings, text, and lane lines, especially surpassing that of the GT, affirming our superior restoration quality. This improvement provides better image restoration for various downstream tasks. Furthermore, compared to other specific as well as unified frameworks, our approach trained with limited labeled data demonstrates greater practical value.

(2) \textit{Objective Evaluation}: Considering that only using subjective evaluation is not comprehensive, we also introduce objective evaluation to evaluate the performance of our method. Objective image quality assessment (IQA) methods can be categorized as full-reference, reduced-reference, and no-reference. Since it is difficult to obtain realistic GT images under adverse weather conditions, and our primary aim is to recover perceptually clearer images, GT images are no longer suitable as a reference. Consequently, the corresponding FR-IQA metrics, such as PSNR and SSIM, are not applicable. Instead, we adopt NR-IQA metrics for image quality evaluation. In this paper, five top-performance metrics are employed for objective evaluation, including PAQ2PIQ~\citep{ying2020patches}, DBCNN~\citep{8576582}, CLIPIQA~\citep{wang2023exploring}, MUSIQ-SPAQ~\citep{ke2021musiq}, and NIQE~\citep{mittal2012making}. In Tables \ref{tab:raindrop}, \ref{tab:rain}, and \ref{tab:snow}, we compare our approach to other fully-supervised approaches, including weather-specific and all-in-one approaches, for these five NR-IQA metrics. It is clear that our method, which employs a semi-supervised learning framework trained with limited labeled data, significantly outperforms other methods overall, even over GT, affirming that the restoration results from our method are visually high-quality.

%--------------------------------------------------------------------------compare raindrop----------------------------------------------------------------%

%--------------------------------------------------------------------------compare raindrop----------------------------------------------------------------%
% %--------------------------------------------------------------------------tabel raindrop------------------------------------------------------------------%
\begin{table*}[htbp!]
\caption{Objective evaluation comparison on \textit{RainDrop} test set in terms of five no-reference IQA (NR-IQA) metrics. The best results are in \textbf{bold} and the second bests are with \underline{underline}.}
\label{tab:raindrop}
\centering
\begin{tabular}{cccccc}
\toprule[1pt]
Method & PAQ2PIQ$\uparrow$ & DBCNN$\uparrow$ & CLIPIQA$\uparrow$ & MUSIQ-SPAQ$\uparrow$  & NIQE$\downarrow$ \\
\midrule[0.8pt]
RaindropAttn~\citep{quan2019deep}       & 72.93            & 67.07            & 0.41            & 72.86            & 20.71       \\
IDT~\citep{xiao2022image}               & \underline{73.38}& 68.55            & 0.42            & \underline{73.92}& \underline{20.26}  \\
UDR-S2Former~\citep{chen2023sparse}     & 73.27            & 68.06            & 0.42            & 73.52            & 20.38      \\
\midrule[0.5pt]
TransWeather~\citep{valanarasu2022transweather} & 71.23 & 60.11 & 0.42 & 71.00 & 20.53 \\
WGWS-Net~\citep{zhu2023learning}        & 73.19            & 68.09            & 0.43            & 73.49            & 20.48        \\
$\text{WeatherDiff}_{64}$~\citep{ozdenizci2023restoring} & 73.13 & \underline{69.59} & 0.46 & 73.72 & 20.27 \\
\midrule[0.5pt]
GT                  & 72.91            & 68.90            & \textbf{0.58}   & 73.16            & 20.27          \\
\midrule[0.5pt]
Ours                & \textbf{74.05}   & \textbf{72.41}   & \underline{0.57}& \textbf{74.88}   & \textbf{19.21}  \\
\bottomrule[1pt]
\end{tabular}
\end{table*}
% \vspace{3.5mm}
%--------------------------------------------------------------------------tabel raindrop------------------------------------------------------------------%
%--------------------------------------------------------------------------compare rain--------------------------------------------------------------------%
% \vspace{2mm}
%--------------------------------------------------------------------------compare rain--------------------------------------------------------------------%
%--------------------------------------------------------------------------tabel rain----------------------------------------------------------------------%
\begin{table*}[htbp!]
\centering
\caption{Objective Evaluation comparison on \textit{Outdoor-rain} test set in terms of five no-reference IQA (NR-IQA) metrics. The best results are in \textbf{bold} and the second bests are with \underline{underline}.}
\begin{tabular}{cccccc}
\toprule[1pt]
Method &PAQ2PIQ$\uparrow$ &DBCNN$\uparrow$ &CLIPIQA$\uparrow$ &MUSIQ-SPAQ$\uparrow$  &NIQE$\downarrow$ \\\midrule[0.8pt]
HRGAN~\citep{li2019heavy}            &72.96&63.02&0.39&73.47&20.10\\
PCNet~\citep{jiang2021rain}            &73.97&61.54&0.36&71.62&20.20    \\
MPRNet~\citep{zamir2021multi}           &74.49&66.80&0.42&73.91&19.86    \\\midrule[0.5pt]
TransWeather~\citep{valanarasu2022transweather}     &73.66&62.35&0.42&72.54&19.54    \\
WGWS-Net~\citep{zhu2023learning}         &74.89&69.46&0.45&74.55&19.74    \\
$\text{WeatherDiff}_{64}$~\citep{ozdenizci2023restoring} &\underline{75.09}&73.39&0.52&\underline{76.02}&19.37  \\\midrule[0.5pt]
GT               &74.92&\underline{74.51}&\textbf{0.60}&75.69&\underline{19.30}\\\midrule[0.5pt]
Ours             &\textbf{76.42}&\textbf{76.31}&\underline{0.57}&\textbf{76.70}&\textbf{18.33}    \\\bottomrule[1pt]
\end{tabular}
\label{tab:rain}
\end{table*}
% \vspace{3.5mm}
%--------------------------------------------------------------------------tabel rain----------------------------------------------------------------------%
%--------------------------------------------------------------------------compare snow--------------------------------------------------------------------%
%--------------------------------------------------------------------------compare snow--------------------------------------------------------------------%
%--------------------------------------------------------------------------tabel snow----------------------------------------------------------------------%
\begin{table*}[htbp!]
\centering
\caption{Objective evaluation comparison on \textit{Snow100K-2000} test set in terms of five no-reference IQA (NR-IQA) metrics. The best results are in \textbf{bold} and the second bests are with \underline{underline}.}
\begin{tabular}{cccccc}
\toprule[1pt]
Method &PAQ2PIQ$\uparrow$ &DBCNN$\uparrow$ &CLIPIQA$\uparrow$ &MUSIQ-SPAQ$\uparrow$  &NIQE$\downarrow$ \\\midrule[0.8pt]
FocalNet~\citep{cui2023focal}            &72.20&55.57&\underline{0.57}&68.39&20.61  \\
SFNet~\citep{cui2023selective}               &72.39&\underline{56.02}&\underline{0.57}&68.62&20.60\\
FSNet~\citep{10310164}              &72.33&55.72&\underline{0.57}&68.61&20.58 
\\\midrule[0.5pt]
TransWeather~\citep{valanarasu2022transweather}        &70.13&51.30&0.51&66.06&20.73  \\
WGWS-Net~\citep{zhu2023learning}            &71.61&52.07&0.53&67.12&20.71   \\
$\text{WeatherDiff}_{64}$~\citep{ozdenizci2023restoring}   &72.54&55.61&\underline{0.57}&68.64&\underline{20.44} \\\midrule[0.5pt]
GT                  &\underline{72.86}&\textbf{56.61}&\textbf{0.59}&\underline{69.22}&20.73  \\\midrule[0.5pt]
Ours                &\textbf{73.20}&55.43&\underline{0.57}&\textbf{69.71}&\textbf{19.43}   \\\bottomrule[1pt]
\end{tabular}
\label{tab:snow}
\end{table*}
% \vspace{3.5mm}
%--------------------------------------------------------------------------tabel snow----------------------------------------------------------------------%
%--------------------------------------------------------------------------compare real_rain---------------------------------------------------------------%
\begin{figure*}[htbp!]
\centering
\captionsetup[subfloat]{labelformat=empty}  % 这将移除子图的标签编号
\captionsetup[subfloat]{captionskip=-1pt}
\subfloat[\scriptsize Input]{%
  \includegraphics[width=0.18\textwidth]{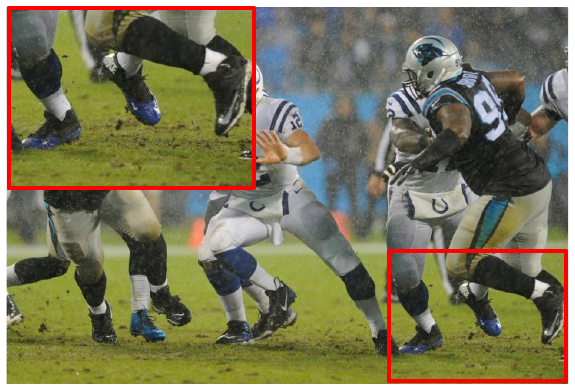}
  \label{fig:input7}
}
\hfill
\subfloat[\scriptsize TransWeather~\citep{valanarasu2022transweather}]{%
  \includegraphics[width=0.18\textwidth]{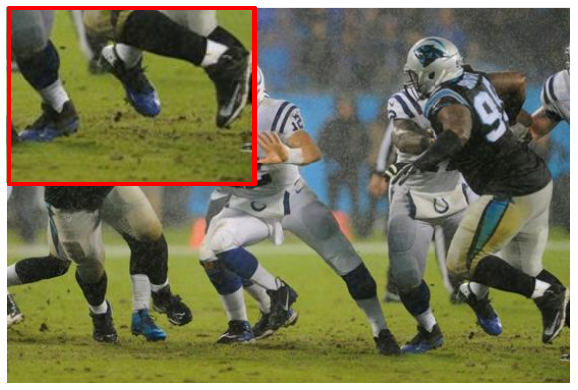}
}
\hfill
\subfloat[\scriptsize WGWS-Net~\citep{zhu2023learning}]{%
  \includegraphics[width=0.18\textwidth]{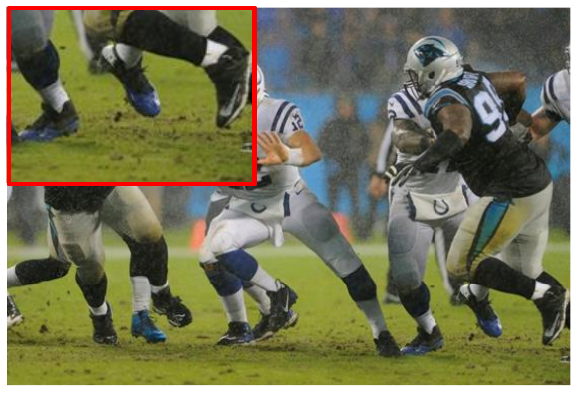}
}
\hfill
\subfloat[\scriptsize WeatherDiff\textsubscript{64}~\citep{ozdenizci2023restoring}]{%
  \includegraphics[width=0.18\textwidth]{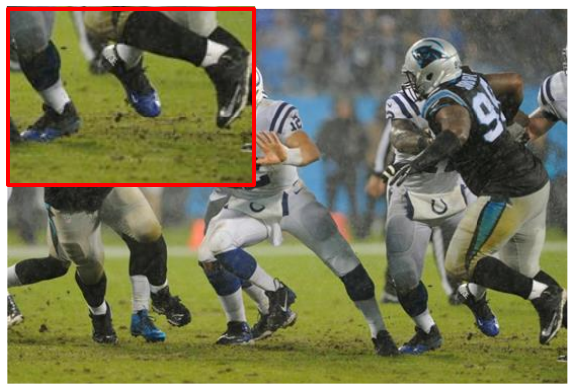}
}
\hfill
\subfloat[\scriptsize Ours]{%
  \includegraphics[width=0.18\textwidth]{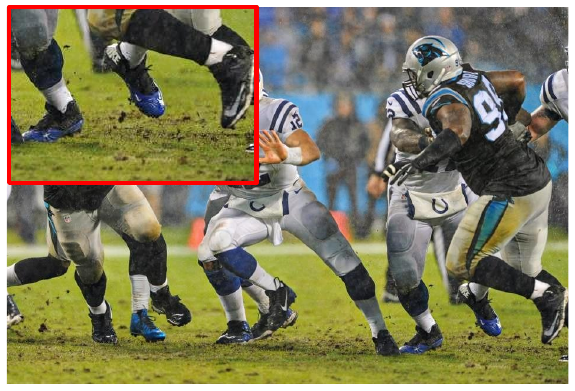}
}
\hfill
\caption{Restoration instances on \textit{IVIPC-DQA} test set by our method and other SOTA fully-supervised competitors. The red box highlights the zoomed-in patch}
\label{fig:compare_real_rain}
\end{figure*}
\vspace{2mm}
%--------------------------------------------------------------------------compare real_rain---------------------------------------------------------------%
%--------------------------------------------------------------------------compare real_snow---------------------------------------------------------------%
\begin{figure*}[htbp!]
\centering
\captionsetup[subfloat]{labelformat=empty}  % 这将移除子图的标签编号
\captionsetup[subfloat]{captionskip=-1pt}
\subfloat[\scriptsize Input]{%
  \includegraphics[width=0.18\textwidth]{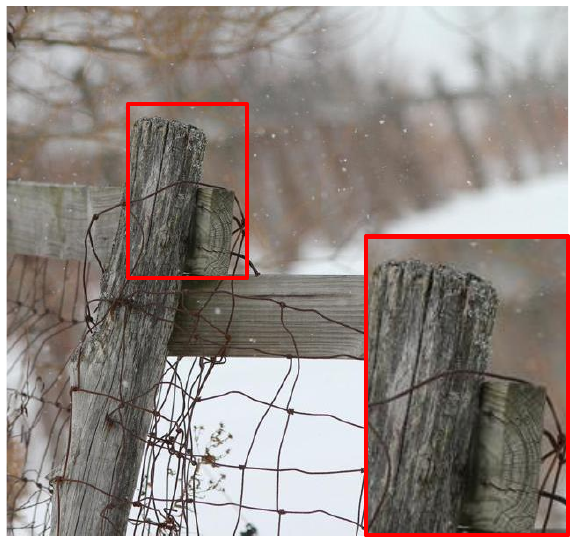}
}
\hfill
\subfloat[\scriptsize TransWeather~\citep{valanarasu2022transweather}]{%
  \includegraphics[width=0.18\textwidth]{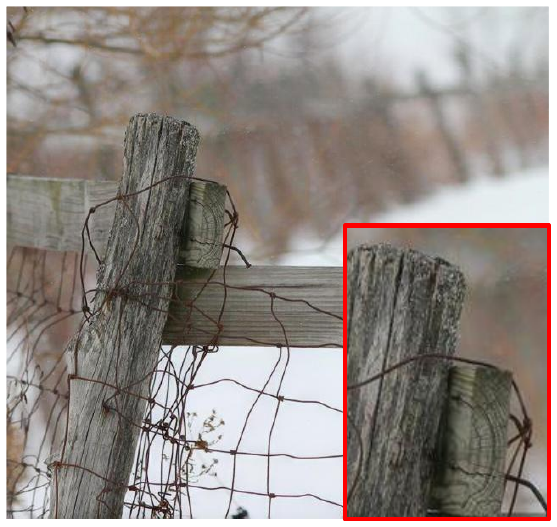}
}
\hfill
\subfloat[\scriptsize WGWS-Net~\citep{zhu2023learning}]{%
  \includegraphics[width=0.18\textwidth]{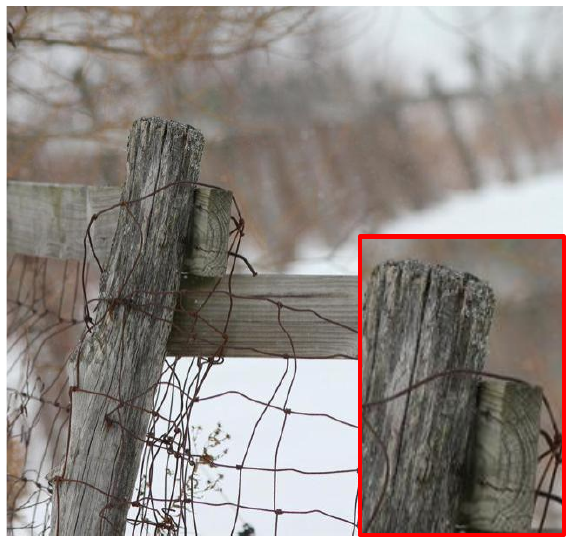}
}
\subfloat[\scriptsize WeatherDiff\textsubscript{64}~\citep{ozdenizci2023restoring}]{%
  \includegraphics[width=0.18\textwidth]{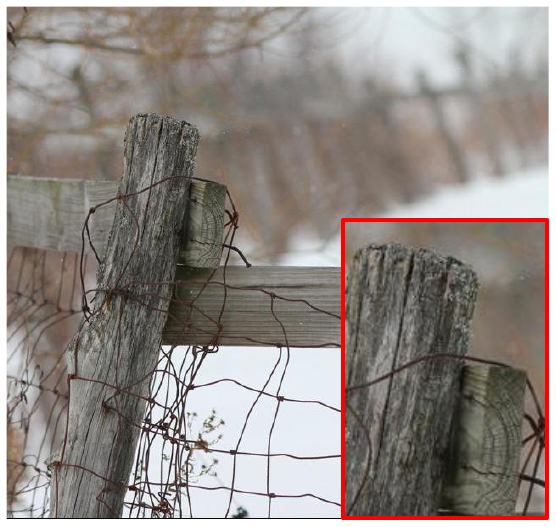}
}
\hfill
\subfloat[\scriptsize Ours]{%
  \includegraphics[width=0.18\textwidth]{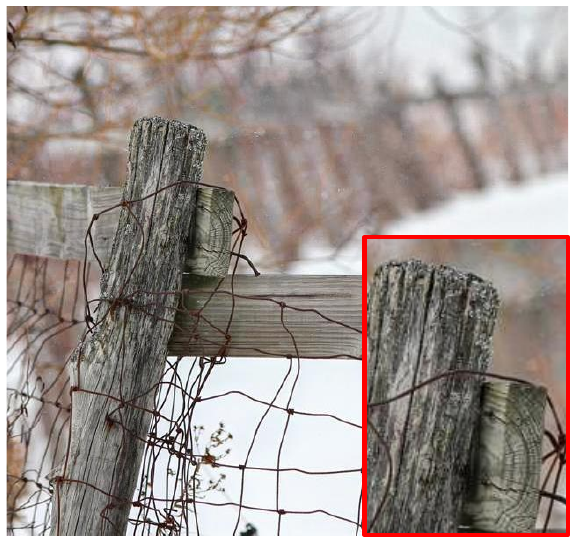}
}
\hfill
\caption{Restoration instances on \textit{Snow100K-Real} test set by our method and other SOTA fully-supervised competitors. The red box highlights the zoomed-in patch}
\label{fig:compare_real_snow}
\end{figure*}
% --------------------------------------------------------------------------compare real_snow-----------------------------------------------------------------%
% --------------------------------------------------------------------------tabel real_rain-----------------------------------------------------------------%

\begin{table*}[htbp!]
\centering
\caption{Objective evaluation comparison on \textit{IVIPC-DQA} test set in terms of five no-reference IQA (NR-IQA) metrics. The best results are in \textbf{bold} and the second bests are with \underline{underline}.}
\begin{tabular}{cccccc}
\toprule[1pt]
Method &PAQ2PIQ$\uparrow$ &DBCNN$\uparrow$ &CLIPIQA$\uparrow$ &MUSIQ-SPAQ$\uparrow$  &NIQE$\downarrow$ \\\midrule[0.5pt]
TransWeather~\citep{valanarasu2022transweather}                     & 71.60 & 49.51                  & \underline{0.51} & 59.64             & 20.54  \\
WGWS-Net~\citep{zhu2023learning}                         & 71.51             & 46.36                  & 0.48             & 57.77             & 20.40   \\
$\text{WeatherDiff}_{64}$~\citep{ozdenizci2023restoring}        & \underline{72.16}             & \underline{52.22}      & \textbf{0.52}    & \underline{59.85} & \underline{20.21}  \\
Ours                             & \textbf{73.41}    & \textbf{56.43}         & 0.49             & \textbf{60.38}    & \textbf{19.14}   \\\bottomrule[1pt]
\end{tabular}
\label{tab:real_metric_rain}
\end{table*}
% \vspace{2mm}
%--------------------------------------------------------------------------tabel real_rain-----------------------------------------------------------------%
%--------------------------------------------------------------------------tabel real_snow-----------------------------------------------------------------%
\begin{table*}[htbp!]
\centering
\caption{Objective evaluation comparison on \textit{Snow100K-Real} test set in terms of five no-reference IQA (NR-IQA) metrics. The best results are in \textbf{bold} and the second bests are with \underline{underline}.}
\begin{tabular}{cccccc}
\toprule[1pt]
Method &PAQ2PIQ$\uparrow$ &DBCNN$\uparrow$ &CLIPIQA$\uparrow$ &MUSIQ-SPAQ$\uparrow$  &NIQE$\downarrow$ \\\midrule[0.5pt]
TransWeather~\citep{valanarasu2022transweather}&             69.44& 50.61&0.55&63.33&20.48\\
WGWS-Net~\citep{zhu2023learning}&                 69.80& 49.92&0.53&62.85&20.57\\
$\text{WeatherDiff}_{64}$~\citep{ozdenizci2023restoring}&\underline{70.49}&\underline{52.88}&\underline{0.56}&\underline{64.22}&\underline{20.46}       \\
Ours&                     \textbf{71.59}& \textbf{53.91}&\textbf{0.57}&\textbf{65.65}&\textbf{19.49}      \\\bottomrule[1pt]
\end{tabular}
\label{tab:real_metric_snow}
\end{table*}
% \vspace{2mm}
%--------------------------------------------------------------------------tabel real_snow-----------------------------------------------------------------%
% --------------------------------------------------------------------------tabel lane_detction-------------------------------------------------------------%
\begin{table*}[htbp!]
\centering
\caption{Performance comparison of lane detection using CLRNet-ResNet-34~\citep{zheng2022clrnet} and ADNet-ResNet-34~\citep{xiao2023adnet} on \textit{CULane} test set: our method versus TKL and WGWS-Net restoration methods. The best results are highlighted in bold.}\label{tab:lane_clrnet}
\begin{tabular}{cccccccc}
\toprule[1pt]
\multirow{2}{*}{Method} & \multicolumn{3}{c}{CLRNet-ResNet-34~\citep{zheng2022clrnet}}  & & \multicolumn{3}{c}{ADNet-ResNet-34~\citep{xiao2023adnet}} \\
\cmidrule(lr){2-4}  \cmidrule(lr){6-8}
&Precision$\uparrow$&Recall$\uparrow$&F1$\uparrow$ & &Precision$\uparrow$&Recall$\uparrow$&F1$\uparrow$ \\
\midrule[0.8pt]
GT  &0.5472 &0.4657 &0.5032 & &0.5220 &0.4495 &0.4831\\
\midrule[0.5pt]
Degraded Image &0.5468 &0.2807 &0.3710 & &0.4936 &0.2394 &0.3224\\
\midrule[0.5pt]
TKL~\citep{chen2022learning}  &0.4131 &0.3153 &0.3576 & &0.4646 &0.2459 &0.3216\\
WGWS-Net~\citep{zhu2023learning} &0.4226 &0.3340 &0.3731 & &0.3983 &0.3244 &0.3576\\
Ours &\textbf{0.5460} &\textbf{0.4337} &\textbf{0.4833} & &\textbf{0.5274} &\textbf{0.4287} &\textbf{0.4730}\\
\bottomrule[1pt]
\end{tabular}
\end{table*}
% \vspace{2mm}
% --------------------------------------------------------------------------tabel lane_detction-------------------------------------------------------------%
%----------------------------------------------------------------abalation rain----------------------------------------------------------------------------%
\begin{table*}[!ht]
\caption{Abalation study of SemiDDM-Weather on \textit{Outdoor-Rain} test set. The best results are in \textbf{bold} and the second bests are with \underline{underline}.}
\begin{center}
{
\begin{tabular}{cccccc}
\toprule[1pt]
Method &PAQ2PIQ$\uparrow$ &DBCNN$\uparrow$ &CLIPIQA$\uparrow$ &MUSIQ-SPAQ$\uparrow$  &NIQE$\downarrow$ \\\midrule[0.8pt]
Ours (baseline)           &\textbf{76.42}&\textbf{76.31}&\textbf{0.57}&\underline{76.70}&18.33  \\\midrule[0.5pt]
w/o aux         &-8.42&-29.25&-0.33&-19.25&+2.64   \\
w/o dynamic warehouse        &-0.08&-2.61&0.0&\textbf{+0.09}&-0.05  \\
w/o contrast    &\underline{-0.06}&\underline{-0.73}&0.0&-0.21&\underline{-0.08}   \\
w/o consistency &-0.26&-4.24&-0.05&-0.10&\textbf{-0.10}   \\
w/o sequence    &-3.13&-14.20&-0.07&-9.64&+3.93  \\\bottomrule[1pt]
\end{tabular}
}
\end{center}
\label{tab:abalation_rain}
\end{table*}
% \vspace{2mm}
%----------------------------------------------------------------abalation rain----------------------------------------------------------------------------%
\subsection{Performance on Real-World Dataset}
To further validate the practicality of our method, we extend our evaluation to real-world data and include the current advanced methods, e.g., TransWeather, WGWS-Net, and $\text{WeatherDiff}_{64}$, for comparison. Figs.~\ref{fig:compare_real_rain} and ~\ref{fig:compare_real_snow} illustrate the deraining instance on the \textit{IVIPC-DQA} test set and the desnowing instance on the \textit{Snow100K-Real} test set, respectively. It is observed that our method not only achieves superior degradation removal but also enhances image clarity, confirming the superiority of our method over other existing methods, such as clearer lawn and sharper tree textures. Tables \ref{tab:real_metric_rain} and ~\ref{tab:real_metric_snow} show the objective evaluation results on the \textit{IVIPC-DQA} test set and the \textit{Snow100K-Real} test set, respectively. Similarly, it can be observed that our method achieves the best results.
\subsection{Application in the Wild}
In this part, we conduct further experiments to verify the practical effect of our method in enhancing the performance of downstream vision applications, such as lane detection. Since TransWeather fails on these downstream tasks, and $\text{WeatherDiff}_{64}$ is quite time-consuming, only TKL~\citep{chen2022learning} and WGWS-Net (all-in-one methods) are included for comparison. We selected models trained on the \textit{CULane}~\citep{pan2018spatial} dataset, including  CLRNet-ResNet-34~\citep{zheng2022clrnet}, and ADNet-ResNet-34~\citep{xiao2023adnet}, to verify the practical effect of our method. The related experimental results are shown in Table \ref{tab:lane_clrnet}. These results clearly demonstrate that, in mitigating performance degradation caused by adverse weather conditions in downstream tasks, our method is not only effective but even superior to fully-supervised methods.
\subsection{Ablation Study}
% 这里有可能以后需要补一段说明：这里不是一般性，以xxx数据集为例。或者再补一个一个数据集。
(1) \textit{Auxiliary Loss of the Generator for Labeled Images Training}: Referring to Table~\ref{tab:abalation_rain}, it can be seen that without the auxiliary loss, the performance of our proposed model consistently degrades significantly over all the five metrics. This verifies that our auxiliary loss design contributes to the overall framework. 

(2) \textit{Pseudo-label Dynamic Warehouse for Unlabeled Images Training}: To assess the impact of the pseudo-label dynamic warehouse on our method, we compare the difference in performance with and without the pseudo-label dynamic warehouse. The results in Table~\ref{tab:abalation_rain} show that the introduction of the pseudo-label dynamic warehouse has improved our performance in general, especially on the DBCNN metric.

(3) \textit{Contrastive Loss for Unlabeled Images Training}: As can be seen from the results in Table \ref{tab:abalation_rain}, removing the contrast loss leads to a decrease in overall performance, which proves that the introduced contrast loss is effective to some extent.

(4) \textit{Method for Constructing Pseudo-label Dynamic Warehouse}: We added a consistency constraint to assess the quality of teachers' network predictions. As can be seen in Table \ref{tab:abalation_rain}, this constraint plays an important role in selecting ``optimal'' pseudo-labels.

(5) \textit{Contribution of Memory Replay Training Strategy}: As can be seen in Table \ref{tab:abalation_rain}, the memory replay training strategy has demonstrated significant effectiveness for our model training. 

\subsection{Model Complexity Analysis}
To comprehensively assess the performance, we compare the efficiency of our method with the other DDM-based all-in-one adverse weather removal method, WeatherDiff\textsubscript{64}~\citep{ozdenizci2023restoring}. The results of model parameters (Params.) and floating-point operations (FLOPs) are reported in Table~\ref{tab:flop}.
It is evident that our method, which uses only half as much labeled data as WeatherDiff\textsubscript{64} yet achieves superior restoration performance, is also more efficient and consumes fewer computational resources.
%---------------------------------------flops---------------------------------------------
\begin{table}[h!]
\caption{Model specifications of WeatherDiff\textsubscript{64} and our method including parameters (M) and FLOPs (GB) on a single GPU for one sample}
\begin{tabular}{cccc}
\toprule[1pt]
Method & Params.$\downarrow$ & FLOPs$\downarrow$ \\\midrule[0.8pt]
WeatherDiff\textsubscript{64} (supervised) & 82.962 & 30.026 \\\midrule[0.5pt]
Ours (semi-supervised) & \textbf{55.654} & \textbf{7.179} \\\bottomrule[1pt]
\end{tabular}
\label{tab:flop}
\end{table}
%---------------------------------------flops---------------------------------------------
\section{Conclusion}
To restore visual high-quality images from adverse weather conditions with limited labeled data, this paper has presented a pioneering semi-supervised learning framework by incorporating a Denoising Diffusion Model into the teacher-student network, termed SemiDDM-Weather. 
By customizing the wavelet diffusion model WaveDiff as the backbone, our SemiDDM-Weather can effectively capture complex many-to-one mapping distributions, thereby achieving perceptually clearer image restoration, even surpassing the ground truth (GT).
Furthermore, to enhance the accuracy of pseudo-labels used for training the student network with unlabeled data, we construct a pseudo-label dynamic warehouse to store the ``optimal'' outputs from the teacher network, determined by the combination of quality assessment and content consistency constraint.
Ablation studies validate the effectiveness of our design. Exhaustive experimental results show that our approach achieves consistently visual high-quality and superior adverse weather removal compared to fully-supervised competitors on both synthetic and real-world datasets. Applications on downstream vision tasks further confirm its practicability. 
\bibliographystyle{elsarticle-num}
\bibliography{Bibliography}
\end{document}